\documentclass{article}

\usepackage{PRIMEarxiv}

\usepackage[utf8]{inputenc} 
\usepackage[T1]{fontenc}    
\usepackage{hyperref}       
\usepackage{url}            
\usepackage{booktabs}       
\usepackage{amsfonts}       
\usepackage{nicefrac}       
\usepackage{microtype}      
\usepackage{lipsum}
\usepackage{fancyhdr}       
\usepackage{graphicx}       
\graphicspath{{media/}}     

\usepackage[numbers]{natbib}
\usepackage{multirow}
\usepackage{tabularx}
\usepackage{makecell}
\newcolumntype{C}{>{\centering\arraybackslash}X}
\usepackage{amsmath}
\title{Community Research Earth Digital Intelligence Twin (CREDIT)
\thanks{\textit{\underline{Citation}}: Schreck, J. et al. Community Research Earth Digital Intelligence Twin (CREDIT).}
\author{
  \textbf{John S.\ Schreck}$^{\spadesuit}$$^{\dagger}$$^{\ddagger}$, 
  \textbf{Yingkai Sha}$^{\spadesuit}$$^{\dagger}$$^{\ddagger}$,
  \textbf{William Chapman}$^{\spadesuit}$$^{\dagger}$$^{\mathsection}$, 
  \textbf{Dhamma Kimpara}$^{\dagger}$,\\ 
  \textbf{Judith Berner}$^{\mathsection}$,
  \textbf{Seth McGinnis}$^{\dagger}$,
  \textbf{Arnold Kazadi}$^{\dagger}$, 
  \textbf{Negin Sobhani}$^{\ddagger}$, 
  \textbf{Ben Kirk}$^{\ddagger}$, 
   \textbf{David John Gagne II}$^{\dagger}$$^{\ddagger}$\\[1em]
  Machine Integration and Learning for Earth Systems (MILES) Group$^{\dagger}$ \\
  Computational and Information Systems (CISL) Lab$^{\ddagger}$ \\
  Climate and Global Dynamics (CGD) Lab$^{\mathsection}$\\[1em]
  NSF National Center for Atmospheric Research \\
  Boulder, Colorado, USA\\[1em]
  \texttt{\{schreck, ksha, wchapman, dgagne\}@ucar.edu}
}}

\begin{document}
\maketitle

\renewcommand{\thefootnote}{\fnsymbol{footnote}}
\footnotetext[0]{\(\spadesuit\) These authors contributed equally to this work.}

\begin{abstract}
Recent advancements in artificial intelligence (AI) numerical weather prediction (NWP) have significantly transformed atmospheric modeling. AI NWP models outperform physics-based systems like the IFS on several global metrics while requiring far fewer computational resources. Despite these successes, existing AI NWP models face limitations related to their training datasets and timestep choices, often leading to artifacts that hinder model performance. To begin to address these challenges, we introduce the Community Research Earth Digital Intelligence Twin (CREDIT) framework, developed at NSF NCAR. CREDIT provides a flexible, scalable, and user-friendly platform for training and deploying AI-based atmospheric models on high-performance computing systems, offering an end-to-end pipeline for data preprocessing, model training, and evaluation that democratizes access to advanced AI NWP. We showcase CREDIT’s capabilities on a new AI NWP model: WXFormer, a novel deterministic vision transformer designed to autoregressively predict atmospheric states while mitigating common AI NWP model pitfalls, such as compounding error growth, through techniques like spectral normalization, padding, and extensive multi-step training. To show the flexibility of CREDIT and to have a state-of-the-art model comparison we train the FUXI architecture within the CREDIT framework. Our results demonstrate that both FuXi and WXFormer, when trained on 6-hourly hybrid sigma-pressure level ERA5, generally outperform IFS HRES on 10-day forecasts, potentially offering significant improvements in efficiency and forecast accuracy. The modular nature of the CREDIT platform enables researchers to experiment with various models, datasets, and scaled training options, fostering collaboration and innovation in the scientific community. 
\end{abstract}

\

\section{Introduction}


Rapid advancements in artificial intelligence (AI) numerical weather prediction (NWP) have shaken the foundations of the meteorological community. Spurred by the release of the WeatherBench framework \cite{Rasp2020weatherbench} on the ERA5 reanalysis dataset \cite{hersbach2020era5}, multiple teams spanning motivated individuals \cite{Keisler2022-mp}, universities \cite{Weyn2020-vc, chen2023fuxi}, tech companies \cite{Kurth2023-tl, Lam2023graphcast, Bi2023pangu}, non-profits \cite{Watt-Meyer2023-wu}, and government agencies \cite{Lang2024-aifs} have developed a variety of AI NWP models that have quickly advanced and surpassed the headline global verification scores of the ECMWF integrated forecast system (IFS) global model. In addition to improved verification scores, the AI NWP models require orders of magnitude fewer computational resources to run than conventional NWP. The combination of improved forecast performance at minimal cost opens the door for a flurry of new possibilities in how we can interact with NWP models, including much larger ensembles, more rapid updates, and potentially improved forecast performance relative to traditional NWP. Unfortunately, these seemingly major advancements come with some caveats that have become more apparent once the meteorological community started meticulously investigating these AI models.

Despite these significant advancements, however, a deeper look into the published AI NWP models reveals common limitations, particularly regarding the data used for training. Most of these models, including the ECMWF IFS NWP model, rely on just five state variables (temperature, u-wind, v-wind, water vapor mixing ratio, and surface pressure), while all other variables are diagnosed from those or are updated solely within parameterizations. Additionally, the IFS vertical coordinate system uses hybrid sigma-pressure levels (i.e., model levels), which follow terrain near the surface and relax to pressure levels aloft. The WeatherBench ERA5 dataset uses pressure level data instead, which works well aloft but intersects with terrain near the surface. ERA5 persists the surface values at pressure levels that are below terrain height with the exceptions of temperature and geopotential height, which are extrapolated \cite{Trenberth1993-rg}. 

The existing AI NWP models are still able to produce successful forecasts, but these data choices may be causing artifacts in the predictions that then have to be addressed by complex choices in terms of architecture, training procedure, time-stepping, and post-processing. Additionally, most of the published AI NWP models use a 6-hour timestep, or in the case of Pangu-Weather \cite{Bi2023pangu}, a mixture of models each with a different timestep to delay the accumulation of regression artifacts. Stepping forward the 1-hour Pangu-Weather model results in dramatic error growth after roughly 12 hours, and the error curve with time does not follow the chaotic error growth pattern expected of physics-based models.

Our approach aims to address some of the deficiencies of current AI NWP models through several key improvements. We utilize a more fit-for-purpose training dataset that better represents the complexities of atmospheric dynamics. This is complemented by a carefully selected set of input variables that capture essential meteorological processes. Additionally, we employ a computationally efficient and scalable neural network architecture, adapted to handle the intricacies of weather and climate prediction across various temporal and spatial scales.

Central to our work is the Community Research Earth Digital Intelligence Twin (CREDIT) framework, developed by the Machine Intelligence Learning for Earth Systems (MILES) group at the NSF National Center for Atmospheric Research (NCAR). While other groups have released model weights and necessary code to run these models, CREDIT aims to provide a comprehensive, end-to-end pipeline. It is designed to scale on standard HPC systems while remaining easily accessible and fully supported. It is fully supported on NSF NCAR's Derecho supercomputer, a globally recognized resource for atmospheric and climate research. This framework represents a significant step towards democratizing access to weather and climate emulation technologies. By providing a robust, user-friendly platform that encompasses the entire modeling process—from data preprocessing to model training and evaluation—CREDIT enables the broader scientific community, including researchers, educators, and enthusiasts, to engage with and contribute to advanced atmospheric modeling without the typical barriers to entry.

To demonstrate the versatility and power of the CREDIT platform, we present two key components in this paper. Firstly, we showcase CREDIT's ability to support and modify existing models from the literature, such as the FuXi model \cite{chen2023fuxi}. This capability allows researchers to build upon and refine established approaches within a standardized framework. Secondly, we introduce WXFormer, a new vision transformer model developed within the CREDIT framework.

WXFormer is the latest advancement in deterministic AI-driven atmospheric modeling, specifically designed to autoregressively predict the state of the atmosphere at a selected time resolution. We present 6-hour intervals in this manuscript as many other models currently do and comment on challenges involved with smaller time steps. WXFormer implements several improvements to mitigate compounding error growth, such as spectral normalization of neural network layers, and integrates physical knowledge into its datasets through static variables like solar radiation at the top of the atmosphere. 

WXFormer incorporates padding techniques to handle the spherical nature of Earth in its global weather simulations. The model employs boundary padding along the map boundaries of [0°-360°] longitude and [-90°-90°] latitude, addressing the challenges posed by polar regions and the dateline. Additionally, spectral normalization is utilized to enhance model stability during training and inference. These design choices enable WXFormer to maintain data continuity and physical consistency across the entire globe, allowing for extended simulation periods.

To effectively handle the spherical nature of the Earth in its simulations, WXFormer employs circular padding along the 0-360° longitude line, wrapping data from one edge to the opposite edge to simulate periodic boundaries. This ensures seamless transitions across the dateline. Additionally, a 180-degree shift is applied to align the data correctly at the poles before padding. The top rows from the North Pole are flipped upside down and added above the original data, while the bottom rows from the South Pole are also flipped and added below. This method guarantees smooth transitions at the poles, respecting the convergence of longitudes and preventing discontinuous seams in the data. These enhancements allow WXFormer to perform simulations over extended time periods, depending on the training protocol and goals.

The selection of the CrossFormer vision transformer \cite{crossformer} as the backbone for WXFormer provides several advantages over vanilla ViT models. Primarily, it facilitates a hierarchical attention scheme, allowing for smaller patch sizes without dramatically increasing the model size or memory footprint. This architecture also supports efficient scaling across multiple GPUs, surpassing the capabilities of graph-based networks and enabling effective management of escalating data volumes and complexity. With WXFormer we aim to strike an optimal balance between attention window size and the number of parameters, thereby optimizing computational resources. Moreover, its design yields faster performance compared to similarly sized models, promoting rapid iterations and facilitating real-time applications in atmospheric modeling.

In this manuscript, we present a comprehensive evaluation of the CREDIT framework and specifically WXFormer's performance compared to the FuXi model, providing multiple metrics that demonstrate the model's fidelity across different timescales, from short-term weather predictions to longer-term atmospheric state projections. We also address the model's shortfalls, acknowledging the challenges that are pervasive in the broader landscape of autoregressive machine-learned atmospheric models. By openly discussing these limitations, we aim to foster a transparent dialogue within the community and pave the way for future improvements in AI-driven atmospheric science.

\section{What is CREDIT?}

The CREDIT framework aims to provide a comprehensive research platform for developing and deploying AI-driven models for the earth system, here focused on the atmosphere. It is built on three core components: access to state-of-the-art datasets, a library of advanced models, and an infrastructure designed for scalable training.

CREDIT's approach to data management is a cornerstone of its functionality. The framework currently provides researchers with curated, high-quality datasets crucial for training accurate atmospheric models, ERA5 and CONUS404 as of this writing, with more additions planned. These datasets are preprocessed and formatted to be readily usable. This feature significantly lowers the barrier to entry for researchers new to the field or those without extensive resources for data acquisition and processing. Advanced users can customize a preprocessing framework within CREDIT. Moreover, CREDIT's data pipeline is designed to be extensible, allowing for the integration of new datasets as they become available.

The framework's model library offers a diverse and growing collection of model architectures. This includes simpler fully convolutional-based models like U-Net and its derivatives, as well as state-of-the-art models such as FuXi, the Swin weather model \cite{willard2024analyzing}, the Spherical Fourier Neural Operator (SFNO) \cite{anima23}, with its application in the AI2 Climate Emulator (ACE) model \cite{watt2023ace}, among others.

CREDIT provides a scalable training infrastructure that leverages a standard High-Performance Computing (HPC) system, Derecho, enabling researchers to take full advantage of available computational resources for training large-scale AI models. This framework facilitates end-to-end training software, providing users a template for efficiently training complex models across multiple GPUs without necessarily needing extensive expertise in parallel computing or HPC environments. By managing many of the intricacies of distributed training, CREDIT will enable scientists to focus on their research goals rather than technical details. Additionally, it supports the creation of customized training recipes, enhancing flexibility and adaptability in model training.

The framework offers a user-friendly interface and documentation aimed at making it accessible to a wide range of users, from experienced climate scientists to students beginning their research journey. As we continue to develop and refine these resources, our long-term vision focuses on fostering community-driven development of CREDIT. We aim to create an open and collaborative environment where researchers at various levels of expertise can contribute to the framework's evolution. This approach seeks to leverage collective expertise to enhance CREDIT's capabilities and ensure it addresses the diverse needs of the atmospheric science research community. As the user base grows, we anticipate that community feedback and contributions will play a crucial role in shaping its features, usability, and overall direction.

Beyond its role as a software framework, CREDIT aspires to be an ecosystem that empowers researchers, educators, and enthusiasts to explore new frontiers in atmospheric, ocean, and land-surface modeling. By providing access to top-tier resources, simplifying technical complexities, and lowering barriers to entry, it aims to accelerate research in AI-driven atmospheric science. This approach has the potential to contribute to advancements in our understanding of weather patterns and climate change, possibly playing a role in deepening our collective knowledge of atmospheric processes on both weather and climate scales.

\section{Data}\label{sec:data}

\subsection{ERA5}\label{sec3}

\begin{table}\label{sec3_tab1}
\centering
\caption{The variables of interest in this study}
\renewcommand{\arraystretch}{1.2}
\begin{tabularx}{\textwidth}
{|c|>{\centering\arraybackslash}X|c|c|c|}
\hline
Type & Variable Name & Short Name & Units & Usage \\ 
\hline
Model level variable     & Zonal Wind                     & U     & $\mathrm{m \cdot s^{-1}}$    & Prognostic \\
Model level variable     & Meridional Wind                & V     & $\mathrm{m \cdot s^{-1}}$    & Prognostic \\
Model level variable     & Air Temperature                & T     & $\mathrm{K}$                 & Prognostic \\
Model level variable     & Specific Humidity              & Q     & $\mathrm{kg \cdot kg^{-1}}$  & Prognostic \\
Single level variable    & Surface Pressure               & SP    & $\mathrm{Pa}$                & Prognostic \\
Single level variable    & 2-Meter Temperature            & t2m   & $\mathrm{K}$                 & Prognostic \\
Single level variable    & Meridional Wind at 500 hPa     & V500  & $\mathrm{m \cdot s^{-1}}$    & Prognostic \\
Single level variable    & Zonal Wind at 500 hPa          & U500  & $\mathrm{m \cdot s^{-1}}$    & Prognostic \\
Single level variable    & Temperature at 500 hPa         & T500  & $\mathrm{K}$                 & Prognostic \\
Single level variable    & Geopotential Height at 500 hPa & Z500  & $\mathrm{m}$                 & Prognostic \\
Single level variable    & Specific Humidity at 500 hPa   & Q500  & $\mathrm{kg \cdot kg^{-1}}$  & Prognostic \\
Invariant variable       & Geopotential at surface        & Z$_{SFC}$ & $\mathrm{m^2 \cdot s^{-2}}$       & Input-only \\
Invariant variable       & Land Sea Mask                  & LSM       & n/a                               & Input-only \\
Forcing variable         & Integrated instantaneous solar irradiance  & $I_s$ & $\mathrm{J \cdot m^{-2}}$ & Input-only \\
\hline
\end{tabularx}
\label{tab:variables}
\end{table}

This study uses the European Centre for Medium-Range Weather Forecasts (ECMWF) Reanalysis version  5 (ERA5; e.g., \cite{hersbach2020era5})) as a foundation for the development of CREDIT models. ERA5 is a global reanalysis dataset that has been widely applied in AI NWP model developments. It provides analysis, forecasts, and ensemble prediction quantities on a wide range of variables that cover the atmosphere, land, and ocean. ERA5 is produced using the Integrated Forecasting System (IFS) version Cy41r2, which incorporates state-of-the-art model physics, core dynamics, and data assimilation at ECMWF.

The ERA5 used in this study was obtained from the National Center for Atmospheric Research (NCAR), Research Data Archive (RDA). The N320 regular Gaussian grid version of ERA5 was applied; it has 1280-by-640 horizontal grids and approximately 0.28$^\circ$ grid spacing. For vertical dimensions, 16 model levels: $\mathrm{\left\{10, 30, 40, 50, \ldots, 90, 95, 100, 105, \ldots, 130, 136, 137\right\}}$ were sliced from the overall 137 IFS model levels. Table \ref{sec3_tab1} summarizes the ERA5 variables that were applied. Prognostic upper-air variables, including air temperature, horizontal wind components, and specific humidity, were selected; their N320 regular Gaussian grid quantities were derived and interpolated from the ERA5 T639 spherical harmonics. For surface variables and 500 hPa variables, they were interpolated from the 0.25$^\circ$ ERA5 to the same N320 regular Gaussian grid. The subsetted ERA5 prognostic variables, as described above, were prepared as 1-hourly and 6-hourly datasets from 1979-2022 by following the validity time convention of ERA5 (i.e., 6-hourly quantities were averaged based on the ``ending time'' of hourly quantities). The hourly and 6-hourly dataset was then divided into three parts, with 1979-2018 used for model training, 2018-2019 used for validation, and 2019-2022 used for result verification.

The ERA5 climatology, as functions of day-of-year and hour-of-day, was used for forecast verification. This study collected the ERA5 climatology from Weatherbench2 \cite{Rasp2020weatherbench}, which was computed from the 30-year period of 1990-2020 using sliding windows of 61 days with Gaussian weighting. The ERA5 climatology is used for verification only. For data pre-processing, the mean and standard deviation of each variable were computed within the training set of 1979-2018 and without sliding windows.

\subsection{Forcing inputs}
Two static forcing variables, land-sea mask and geopotential at the surface, were selected from ERA5. Land-sea mask is a float ranging from 0 to 1 indicating the proportion of each grid cell containing water; geopotential at the surface is a measure of terrain height.

Instantaneous solar irradiance ($I_s$) is an estimation of solar power as energy flux that the earth receives at the top of the atmosphere on a specific time and location. $I_s$ is used as the forcing input of CREDIT models.

Given solar constant ($G_{\mathrm{SC}}$), $I_s$ is derived as follows:

\begin{equation}
I_s = \max\left\{G_{\mathrm{SC}} \cdot \left(\frac{1}{d}\right)^2 \cos\left(\theta\right), 0\right\}
\end{equation}

\noindent
Where $d$ is the earth-sun distance in astronomical units and $\theta$ is the zenith angle. $I_s$ is non-negative and has spatiotemporal variations, with $G_{\mathrm{SC}}$ and $d$ vary by time, and $\theta$ varies by time, latitude, longitude, and altitude. 

The estimation of $G_{\mathrm{SC}}$ is consistent with IFS-Cy41r2. For 1979-1995, $G_{\mathrm{SC}}$ was obtained from observational values. For 1996 and onwards, the 13-year cycle from 1983-1995 was repeated. The $G_{\mathrm{SC}}$ of a specific date and time was interpolated linearly between the two years using Gregorian calendar as reference time (i.e., 365.2425 days per year). 

$d$ is computed using the NREL-SPA algorithm as described in \cite{reda2003solar}. In practice, $I_s$ with a unit of $\mathrm{W\cdot m^{-2}}$ was calculated on every minute as $\mathrm{J\cdot m^{-2}}$ and accumulated to 1-hour and 6-hour intervals to match with pre-processed ERA5 datasets.

\subsection{Data Preprocessing}

All ERA5 variables and forcing inputs, except the land-sea mask, were standardized using z-scores. Additionally, residual normalization, as adopted from \cite{watt2023ace}, was applied to the standardized variables.

For each variable (denoted as $T$), the tendency of its standardized version is calculated:

\begin{equation}
\Delta T' = \left\{ T'(t=1) - T'(t=0), \ldots, T'(t+1) - T'(t) \right\}, \quad T' = \frac{T - \mu(T)}{\sigma(T)}
\end{equation}

\noindent
where $\mu$ and $\sigma$ represent the mean and standard deviation, respectively, and $t$ denotes time in either 6-hour intervals.

The residual normalization coefficient ($\xi$) is the standard deviation of $\Delta T'$, re-scaled across all variables using geometric mean. For data pre-processing, $\xi$ is applied together with $\sigma$:

\begin{equation}
T'' = \frac{T-\mu(T)}{\xi(T)\sigma(T)}, \quad \xi(T) = \frac{\sigma(\Delta T')}{\mathrm{gmean}[\sigma(T'_0), \sigma(T'_1), \ldots, \sigma(T'_N)]}
\end{equation}

\noindent
where $T''$ is the pre-processed variable after residual normalization and $N$ represents all pre-processed variables. Note that the calculation is performed independently at each vertical level for upper-air variables.

The residual normalization described above functions similarly to the per-variable-level inverse variance loss described in \cite{Lam2023graphcast}. It re-scales variables based on the amplitude of their temporal evolution, and thus, makes model training converge faster. Variables that vary more strongly over space than over time, such as surface pressure and 500 hPa geopotential height, were found to receive larger performance gains from residual normalization.

\section{Methods}

\subsection{WXFormer}

\begin{figure}[t!]
    \centering
    \includegraphics[width=\columnwidth]{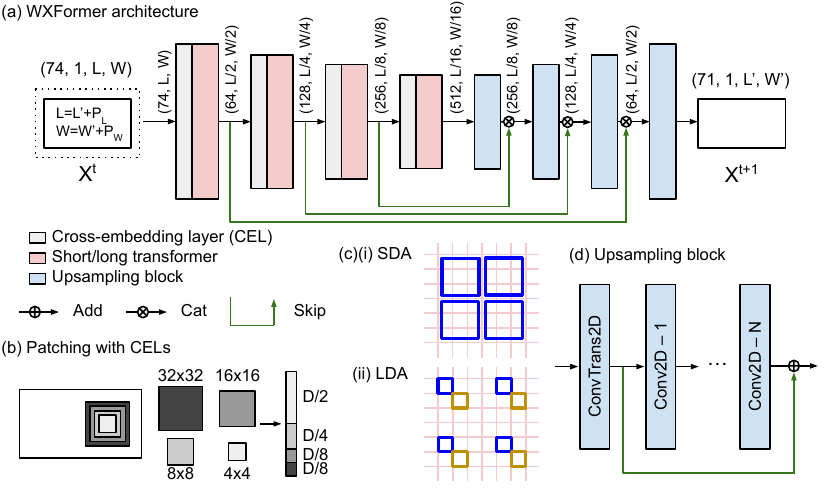}
    \caption{(a) The WXFormer architecture consists of encoding stages using a CrossFormer backbone and decoding stages with hierarchical transpose convolutional layers, with skip connections for improved feature flow. (b) The CEL captures multi-scale features using four convolutional kernels. The LSDA mechanism includes (c)(i) SDA for local interactions and (c)(ii) LDA for global dependencies. (d) The decoder component employs convolutional upsampling blocks with skip connections to progressively increase feature map resolution and maintain spatial information.}
    \label{fig:model}
\end{figure}

WXFormer is a hybrid architecture developed at NCAR that consists of encoding stages using a CrossFormer backbone \cite{crossformer} and decoding stages with hierarchical transpose convolutional layers. Skip connections were assigned from encoders to decoders similar to that of U-net \cite{unet} having overall a pyramid structure. Its overall design is illustrated in Figure~\ref{fig:model}(a). WXFormer leverages the multi-scale feature processing and long-range dependency modeling of CrossFormer backbones, while incorporating efficient feature processing and detail preservation from the pyramid structure. CrossFormer, the transformer basis of WXFormer, has demonstrated similar performance compared to other comparable vision transformers such as Swin-Transformers, already used in AI-NWP approaches \cite{chen2023fuxi, willard2024analyzing} where it forms the transformer foundation of the FuXi weather model \cite{chen2023fuxi}. Similar to models incorporating Swin, this architectural choice potentially offers advantages in terms of feature representation and model efficiency. The model takes as input the state of the atmosphere at times $i$ and is tasked with predicting the state at $i+1$ at a one-hour time step. 

The Cross-scale Embedding Layer (CEL), depicted in Figure~\ref{fig:model}(b), forms the foundation of CrossFormer and hence WXFormer. CEL employs a multi-kernel approach, utilizing four convolutional kernels (4-by-4, 8-by-8, 16-by-16, and 32-by-32) in the initial stage, all with a 4-by-4 stride \cite{crossformer}. This multi-scale sampling captures both fine-grained and coarse-grained image features. To optimize computational efficiency, CEL implements a dimension allocation strategy, assigning fewer dimensions to larger kernels and more to smaller ones. For a 128-dimensional embedding, this might translate to 64 dimensions for 4-by-4, 32 for 8-by-8, and 16 each for 16-by-16 and 32-by-32 kernels.

The Long Short Distance Attention (LSDA) mechanism, crucial to WXFormer, is divided into Short Distance Attention (SDA) and Long Distance Attention (LDA), as shown in Figure~\ref{fig:model}(c). SDA operates on local GxG neighborhoods, while LDA facilitates long-range interactions by sampling embeddings at a fixed interval $I$. This approach bears similarities to axial attention, particularly in its global attention scheme \cite{axial_att}. The global attention is performed across the windowing dimension, reducing computational complexity in a manner analogous to axial attention. This dual attention mechanism enables efficient processing of both local and global contextual information, reducing computational complexity from $O(S^4)$ to $O(S^2G^2)$, where $S$ is input size and $G << S$.

To handle variable-sized inputs, WXFormer incorporates a Dynamic Position Bias (DPB) module. The DPB is a significant improvement over traditional fixed-size relative position bias matrices. It generates relative position bias dynamically using an MLP-based structure comprising three fully-connected layers. The input to this module is a two-dimensional vector ($\Delta$x, $\Delta$y) representing the coordinate distance between two embeddings. The first two layers of the DPB have a dimension of D/4, where D is the embedding dimension, and employ Layer Normalization followed by ReLU activation. The final layer reduces the output to a single scalar value, which serves as the position bias. This dynamic approach allows WXFormer to adapt to different image and group sizes without the limitations of fixed-size bias matrices. Moreover, the DPB is trainable and optimized end-to-end with the rest of the model, enabling it to learn complex positional relationships that may vary across different scales and regions of the input.

The principle decoder component of WXFormer, shown in Figure~\ref{fig:model}(d), incorporates convolutional upsampling blocks with skip connections from corresponding transformer layers. Each block comprises a ConvTranspose2D layer followed by a Conv2D layer, with a residual connection between the ConvTranspose2D output and the final Conv2D output. The ConvTranspose2D layer increases spatial dimensions of feature maps, while the Conv2D layer refines upsampled features and adjusts channel counts. This architecture block progressively increases feature map resolution while maintaining fine-grained spatial information, enabling accurate reconstruction of detailed outputs.

Spectral normalization \cite{spec_norm} and boundary padding along the map boundaries of [0°-360°] and [-90°-90°] were incorporated as additional model designs of WxFormer. Spectral normalization has proven effective in preventing model instabilities during training and avoiding mode collapse in inference (e.g.,\cite{spec_norm, lin2021spectral}). The boundary padding technique is crucial for handling polar and dateline information. Specifically, for polar regions, a 180-degree shift is applied, with top rows from the North Pole flipped and added above the original data, and bottom rows from the South Pole flipped and added below. For the dateline, circular padding is applied along the 0-360° longitude line, wrapping data from one edge to the opposite. This comprehensive approach ensures seamless transitions across all global boundaries, maintaining the physical integrity of the simulations. Finally, our AI NWP models can sometimes produce negative specific humidity values in their raw output. To address this issue, we introduced a non-negative correction on specific humidity as part of the model design. Specifically, any de-normalized specific humidity values that fall below 1E-8 are hard-corrected to 1E-8.

\subsection{Baseline method: FuXi}

FuXi, a state-of-the-art AI NWP model, was selected as the AI NWP model baseline. The implementation of FuXi baseline follows its original design as in \cite{chen2023fuxi} but with reduced model sizes (hereafter, ``FuXi baseline'' or ``FuXi'').

The modification of the FuXi baseline was driven by several considerations. First, the WXFormer design prioritized the balance between model size and performance, resulting in a relatively small model. To ensure a fair comparison, it was reasonable to re-scale FuXi to a size that is comparable to WXFormer. Second, using N320 regular Gaussian grid reduced the input tensor sizes, making it feasible to reduce the FuXi model sizes without downgrading its performance. Different from \cite{chen2023fuxi}, which developed three cascaded FuXi models for 0-15 days of forecasts, this study develops only one FuXi baseline for 0-10 days.

The re-scaling of the FuXi model was guided by \cite{willard2024analyzing}, which explored the effectiveness of hyperparameters on the design of a Swin-Transformer-based AI NWP model. Based on \cite{willard2024analyzing}, window size and patch size are the most important hyperparameters that can impact model performance; larger embedding dimensions are generally beneficial, and model depth has the least impact. With these findings, we preserved the original window and patch sizes from FuXi, slightly reduced its embedding dimensions from 1536 to 1024, and largely reduced its depth from 48 to 16 Swin-Transformer blocks. Spectral normalization, boundary padding, and non-negative correction on specific humidity, as in WXFormer, were added to the re-scaled FuXi baseline to improve its stability and performance.

The FuXi baseline serves as a proof of methodology. By benchmarking WXFormer against a successful AI NWP model, its overall level of effectiveness can be evaluated. In addition, the results and comparisons with the FuXi baseline will demonstrate the efficacy of the CREDIT framework, which supports flexible re-scaling and training of state-of-the-art AI NWP models based on user demand.

\subsection{Model training}
\label{model_training}

The training of AI NWP models was carried out in two stages. The first stage is single-step pre-training, where models were trained to predict the state of the atmosphere for the next time step. The second stage is multi-step fine-tuning, where the models were trained to predict multiple consecutive future states autoregressively, and the loss from each forecasted state was accumulated and used to optimize the model weights. Single-step pre-training provides the basic information for AI NWP models to understand the evolution of atmospheric variables, whereas multi-step fine-tuning aims to enhance model performance and stability for longer forecast lead times, but may also introduce problems such as overly smoothed outputs and decreased ensemble spread (e.g., \cite{price2023gencast, brenowitz2024practical}). The combination of the two stages has proven effective in many AI NWP model studies (e.g., \cite{chen2023fuxi,Lam2023graphcast,willard2024analyzing, siddiqui2024exploring}). 

For the single-step pre-training, cosine-annealing schedules were applied with a set of initial learning rates: $\left\{\mathrm{1E-3}, \mathrm{1E-4}, \mathrm{1E-5}\right\}$. When the higher initial learning rate training is early-stopped based on the validation set performance, the next lower learning rate training will start until the 1E-5 scheduler is early-stopped or a total number of 70 epochs reached.

The multi-step fine-tuning was conducted on 06-48 forecast lead times (i.e., fixed 8 autoregressive steps for 6-hourly models, and fixed 24 steps for the hourly WXFormer) were accumulated to update the weights. The same cosine-annealing schedule was used with initial learning rates of $\left\{\mathrm{1E-4}, \mathrm{1E-5}, \mathrm{1E-6}\right\}$. For 6-hourly models, a total number of 15 epochs was trained. This approach is similar to that of \cite{willard2024analyzing}. For 1-hourly WXFormer, due to resource constraints, its 24-step fine-tuning was not conducted on full epochs, but 35 epochs with shuffled 100 batches per epoch. 

Both stages used latitude-weighted mean squared error as loss functions, the AdamW optimizer \cite{loshchilov2017decoupled}, and batch sizes of 32. The training was conducted on 32 NVIDIA A100 GPUs, each with 40 GB memory, using Pytorch \cite{paszke2019pytorch}. Several technical strategies, including fully sharded data parallel \cite{zhao2023pytorch} specifically focused on the attention layers in the models, activation checkpointing, and loss gradient accumulation with delayed backpropagation, were applied to reduce memory footprints. For 6-hourly models, the time cost is 10-hour per single-step epoch and 18-hour per multi-step epoch. For hourly WXFormer, the time cost is 16-hour per single-step epoch and 4-hour per 100 training batches.

After training, the 6-hourly WXFormer and 6-hourly FuXi models produced results on 6-hourly forecast lead times up to day-10 (i.e., 006, 012, ..., 240 hour forecasts) using the ERA5 dataset as initializations. In contrast, the hourly WXFormer model produced results on hourly forecast lead times up to day-5.

\subsection{Baseline method: IFS-HRES}

The ECMWF Integrated Forecast System (IFS) features a high-resolution (HRES) configuration with 0.1$^\circ$ grid spacing. IFS-HRES is recognized as the best global, operational, medium-range weather forecast. Its initial conditions were estimated every 6 hours using an ensemble 4D-Var system with forecast inputs from the previous assimilation cycle and observations within a 3-hour time window. The IFS-HRES forecasts are operated four times per day. Its 00 and 12Z initialization would go up to day 10, whereas the 06 and 18Z initializations are available for up to day 4.

In this study, the 00 and 12Z IFS HRES baselines were obtained from the Weatherbench 2 data archive \cite{rasp2024weatherbench}, and interpolated to the N320 regular Gaussian grid for verifications. IFS-HRES provides a comparable NWP baseline for AI NWP models that were initialized from and evaluated against ERA5. However, it is important to note that due to the additional interpolation step and the ongoing improvements of IFS, the operational model is expected to be more skillful. Therefore, the IFS HRES verification results of this study should be interpreted with caution.

\subsection{Verfication methods}

\subsubsection{Verification scores}

Pre-trained 6-hourly models and hourly WXFormer are verified in a three-year period of 2019-2022 with ERA5 initializations 00 and 12 UTC daily. The 6-hourly models produced iterative forecasts for up to day-10 (i.e., 40 forecast steps), whereas the hourly WXFormer produced iterative forecasts for up to day-5 (i.e., 120 forecast steps).

Verification scores of Root Mean Square Error (RMSE) and Anomaly Correlation Coefficient (ACC) were calculated as in \cite{rasp2024weatherbench} and for every 6 hours, for both hourly WXFormer and 6-hourly models, up to day 10. Note that all verified variables are instantaneous, which means only the 6-hourly ERA5 dataset and 6-hourly climatology are used.

RMSE measures the average magnitude of the forecast errors. Given forecast $F$ and the ERA5 verification target $O$, RMSE as functions of initialization time ($t_i$) and forecast lead time ($t_l$) is defined as follows:

\begin{equation}
\mathrm{RMSE}(t_i, t_l) = \sqrt{\frac{1}{N_{\phi} N_{\lambda}} \sum_{i=1}^{N_{\phi}}{ \sum_{j=1}^{N_{\lambda}} {\left\{w(i) \left[F(t_i, t_l, i, j) - O(t_i + t_l, i, j)\right]^2\right\}}}}
\end{equation}

\noindent
Where $N_{\phi}$ and $N_{\lambda}$ are the number of latitude ($\phi$) and longitude ($\lambda$) grid cells, respectively. $w\left(i\right) = \cos(\phi_i)$ is the latitude weighting coefficient. $t_i + t_l$ is the derivation of observational time based on initialization and forecast lead times. $\mathrm{RMSE}\left(t_i, t_l\right)$ was computed on target variables and bootstrapped on the $t_i$ dimension to produce verification results as functions of $t_l$.

ACC is the Pearson correlation coefficient between the anomalies of $F$ and $O$:

\begin{equation}
\mathrm{ACC}(t_i,t_l) = \frac{\overline{\mathrm{Cov}}\left[F'(t_i,t_l,i,j),O'(t_i+t_l,i,j)\right]}{\sqrt{\overline{\mathrm{Var}}\left[F'(t_i,t_l,i,j)\right]} \sqrt{\overline{\mathrm{Var}}\left[O'(t_i+t_l,i,j)\right]}}
\end{equation}

\noindent
Where $F'=F-C$ and $O'=O-C$ are anomalies computed from the ERA5 climatology ($C$). $\overline{\mathrm{Cov}}$ and $\overline{\mathrm{Var}}$ are the latitude-weighted average of covariance and variance:

\begin{equation}
\begin{aligned}
\overline{\mathrm{Cov}}\left(F', O'\right) &= \frac{1}{N_{\phi} N_{\lambda}}\sum_{i=1}^{N_{\phi}} \sum_{j=1}^{N_{\lambda}} \left[w(i) F'\left(t_i,t_l,i,j\right) O'\left(t_i+t_l,i,j\right)\right] \\
\overline{\mathrm{Var}}\left(F'\right) &= \frac{1}{N_{\phi} N_{\lambda}}\sum_{i=1}^{N_{\phi}} \sum_{j=1}^{N_{\lambda}} \left[w(i) F'\left(t_i,t_l,i,j\right)^2 \right] \\
\overline{\mathrm{Var}}\left(O'\right) &= \frac{1}{N_{\phi} N_{\lambda}}\sum_{i=1}^{N_{\phi}} \sum_{j=1}^{N_{\lambda}} \left[w(i) O'\left(t_i+t_l,i,j\right)^2 \right]
\end{aligned}
\end{equation}

\noindent
Similar to RMSE, $\mathrm{ACC}\left(t_i, t_l\right)$ was bootstrapped to produce verification results as functions of $t_l$. 

Perfect forecasts have $\mathrm{ACC}=1.0$. Given $F'$ and $O'$ with their mean values close to 0, ACC and Mean Squared Error (MSE) have the following relationships \cite{murphy1989skill}:

\begin{equation}
\frac{\mathrm{MSE}\left(F,O\right)}{\mathrm{MSE}\left(C,O\right)} \approx 2\mathrm{ACC}\left(F',O'\right) - 1
\end{equation}

\noindent
That said, RMSE and ACC share the same point-to-point verification aspects, with RMSE focusing on the absolute difference and ACC focusing on the forecast skill gains relative to a climatology reference.

\subsubsection{Global energy spectrum}

The verification of the global energy spectrum is computed using spherical harmonic transforms. For a given forecasted or analyzed field $F\left(\phi, \lambda\right)$, it can be represented using spherical harmonic functions $Y\left(\phi, \lambda\right)$ as orthonormal basis and spherical harmonic coefficients ($a$):

\begin{equation}
F\left(\phi, \lambda\right) = \sum_{l=0}^{l_{\mathrm{max}}}{\sum_{m=-l}^l{a_l^m Y_l^m\left(\phi, \lambda\right)}}
\end{equation}

\noindent
Where degree $l$ represents the total angular frequency of $Y$. $m$ is the zonal wave number. The energy spectrum of $F$ at a given $m$ is the sum of magnitudes of $a$ in all degrees with $l\geq m$:

\begin{equation}
P\left(m\right) = \sum_{l\geq m}{\left\|a_l^{m}\right\|^2}
\end{equation}

The kinetic energy ($\mathrm{m^2\cdot s^{-2}}$) and potential temperature energy ($\mathrm{K^2}$) spectrum on 500 hPa pressure level were computed and as functions of $m$, $t_i$, and $t_l$. The result is averaged on $t_i$. Comparing $P\left(m\right)$ on forecasts and the ERA5 target, the ability of weather prediction models to represent the energy transfer across scales can be verified. In addition, the energy spectrum provides a measure of the effective resolution of AI NWP models, which helps identify the smoothing effect of neural-network-based computations and model training.

\subsection{Spatial correlation}

Spatial correlation coefficients were computed for ERA5 and AI NWP model outputs on upper-air variables (Table \ref{sec3_tab1}) at all model levels.

For each $t_i$ and $t_l$ (or $t_i+t_l$ for the ERA5) and model level, upper-air variables were flattened to one-dimensional vectors. Pearson correlation coefficients were then computed between variables on their flattened spatial locations. Spatial correlation coefficients on each $t_i$ were computed separately and averaged over the entire verification period to examine the ability of AI NWP models to preserve the relationships between variables across levels and locations.

\section{Results}

\subsection{Verification scores}

\begin{figure}
    \centering
    \includegraphics[width=\columnwidth]{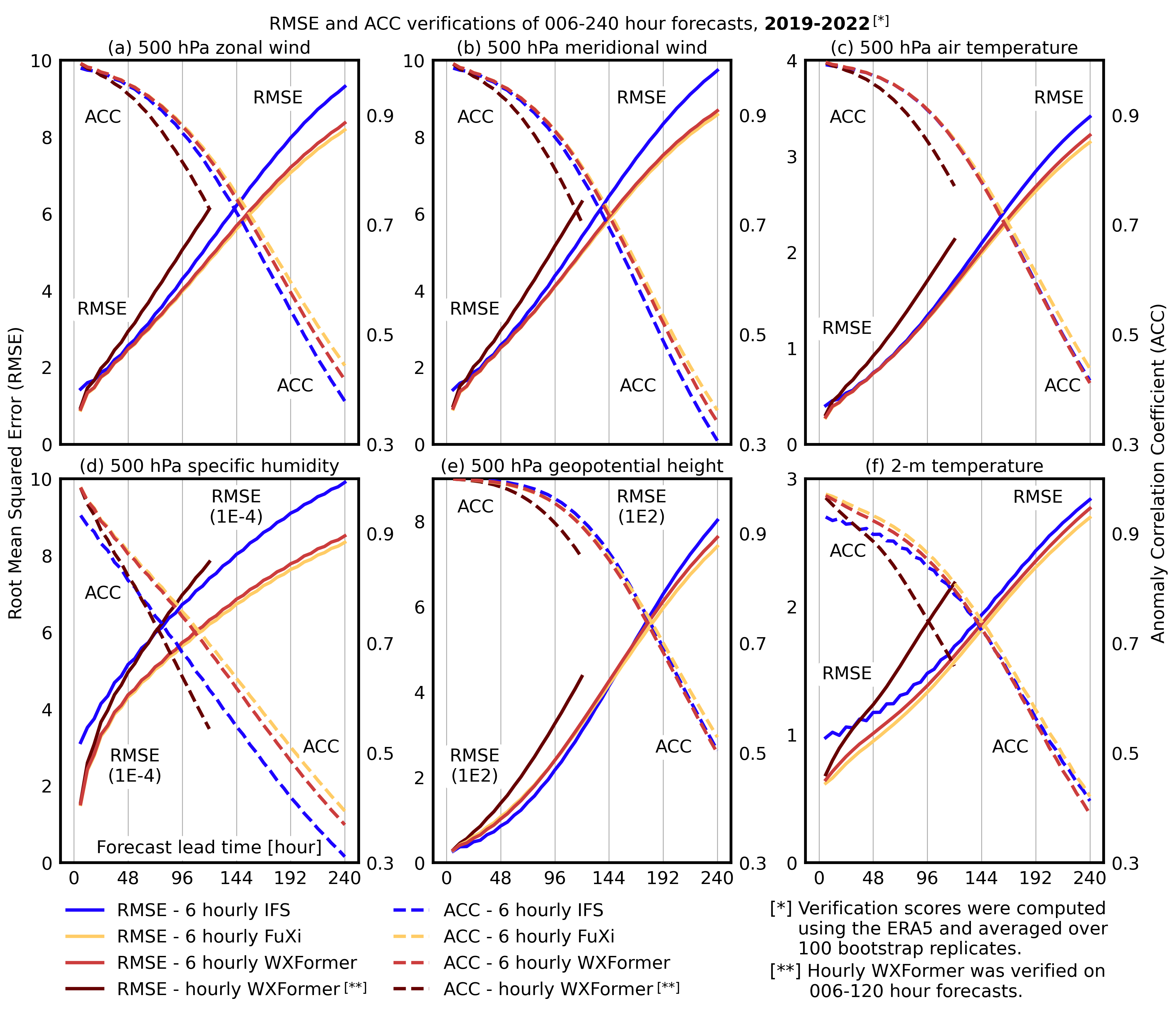}
    \caption{RMSE (solid) and ACC (dashed) for 6-hourly WXFormer (red), 6-hourly FuXi (orange), hourly WXFormer (dark red), and IFS (blue) for different atmospheric variables during a 10-day forecast period (5-day period for the hourly WXFormer). Panels (a-f) show the forecast performance of 500 hPa zonal wind, 500 hPa meridional wind, 500 hPa air temperature, 500 hPa specific humidity, 500 hPa geopotential height, and 2-meter temperature, respectively.}
    \label{fig2}
\end{figure}


In Figure~\ref{fig2}, panels (a) through (f), illustrate the RMSE and ACC for various atmospheric variables for a 10-day forecast. WXFormer (orange) and FuXi (red) are compared against IFS (blue). WXFormer and FuXi demonstrate competitive or better performance compared to IFS at extended lead times, particularly for variables like 500 hPa u- and v-wind (a and b) and specific humidity (d), where they exhibit lower RMSE and comparable or better ACC, especially from 144 to 240 hours. The only standout is geopotential height in which IFS does slightly better than both models until about the five day mark when the trend reverses. Notably, while FuXi was not originally developed by us, we introduced spectral normalization, boundary padding, and non-negative correction on specific humidity to it, resulting in a more efficient version that does not require the cascading approach used in the original paper. These enhancements, along with the residual loss mechanism in both WXFormer and FuXi, contribute to a stable and robust performance across lead times, allowing the models to handle complex atmospheric dynamics effectively.

\subsection{Verification of energy spectrum}

\begin{figure}
    \centering
    \includegraphics[width=\columnwidth]{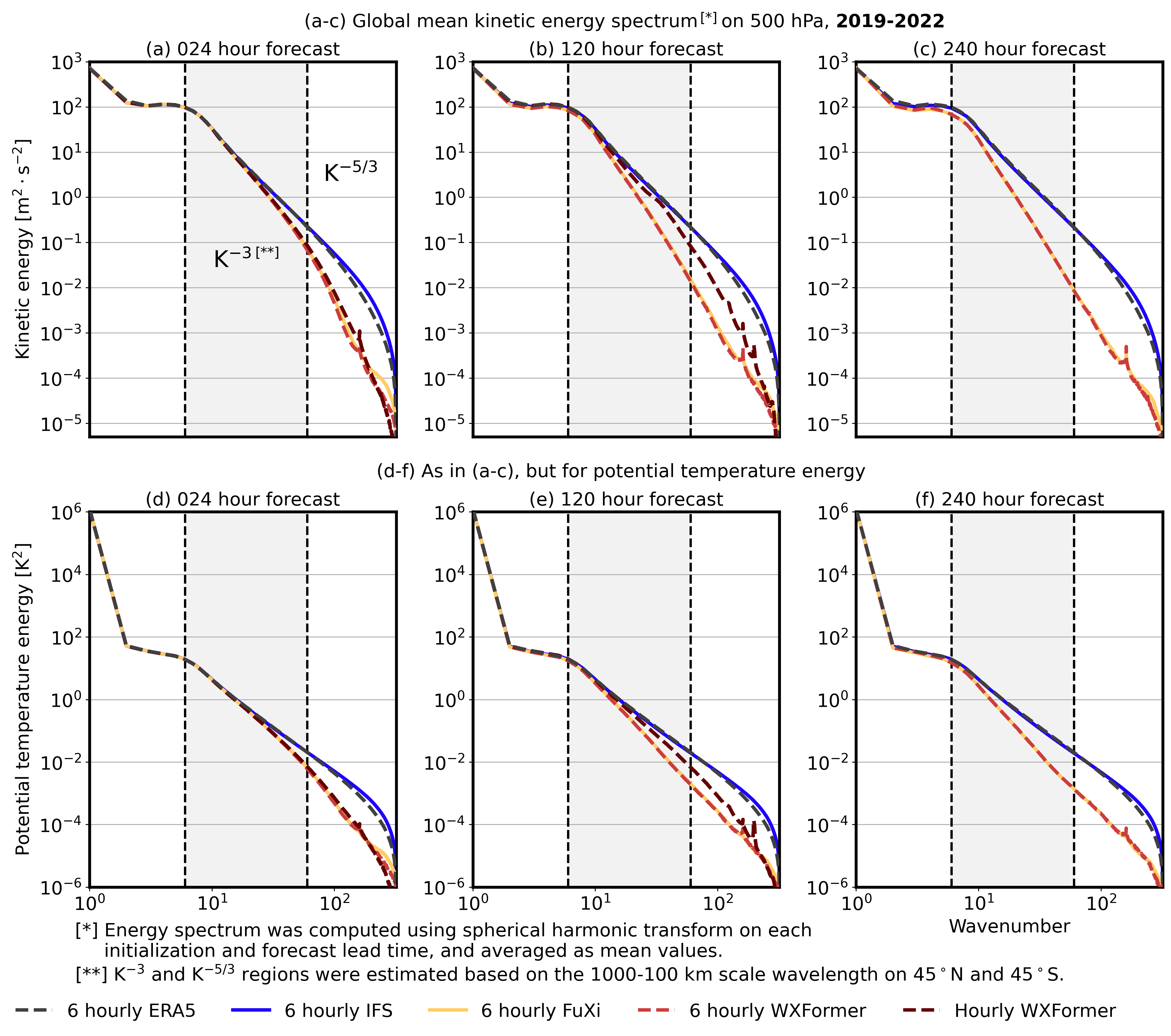}
    \caption{Global mean kinetic energy and potential temperature energy spectra for 6-hourly WXFormer (red), 6-hourly FuXi (orange), hourly WXFormer (dark red), and IFS (blue) for forecasts at different lead times at 500 hPa. The dashed black line shows ERA5. Panels (a-c) show the kinetic energy spectra for forecast lead times of 18-24 hours, 114-120 hours, and 234-240 hours, respectively. Panels (d-f) show the corresponding potential temperature energy spectra. Hourly WXFormer results were presented on (a), (b) and (d), (e).}
    \label{fig3}
\end{figure}

Figure~\ref{fig3} shows the model spectral energy over different lead times for kinetic energy (a-c) and potential temperature energy (d-f) at 500 hPa. The energy spectra are plotted for forecast intervals of 18-24 hours, 114-120 hours, and 234-240 hours, revealing how the models evolve during the forecast. Both WXFormer and FuXi show a loss of spectral resolution as lead time increases, particularly evident in (c) and (f). The spectral slopes smooth with increased forecast lead-time, indicating increased energy dissipation and a diminished ability to capture small-scale features. In contrast, IFS exhibits a retains spectral energy consistent with the ERA5 product at smaller scales over longer forecasts, suggesting a more effective preservation of fine-scale atmospheric features throughout the forecast period.


\subsection{Spatial correlation analysis}

\begin{figure}
    \centering
    \includegraphics[width=\columnwidth]{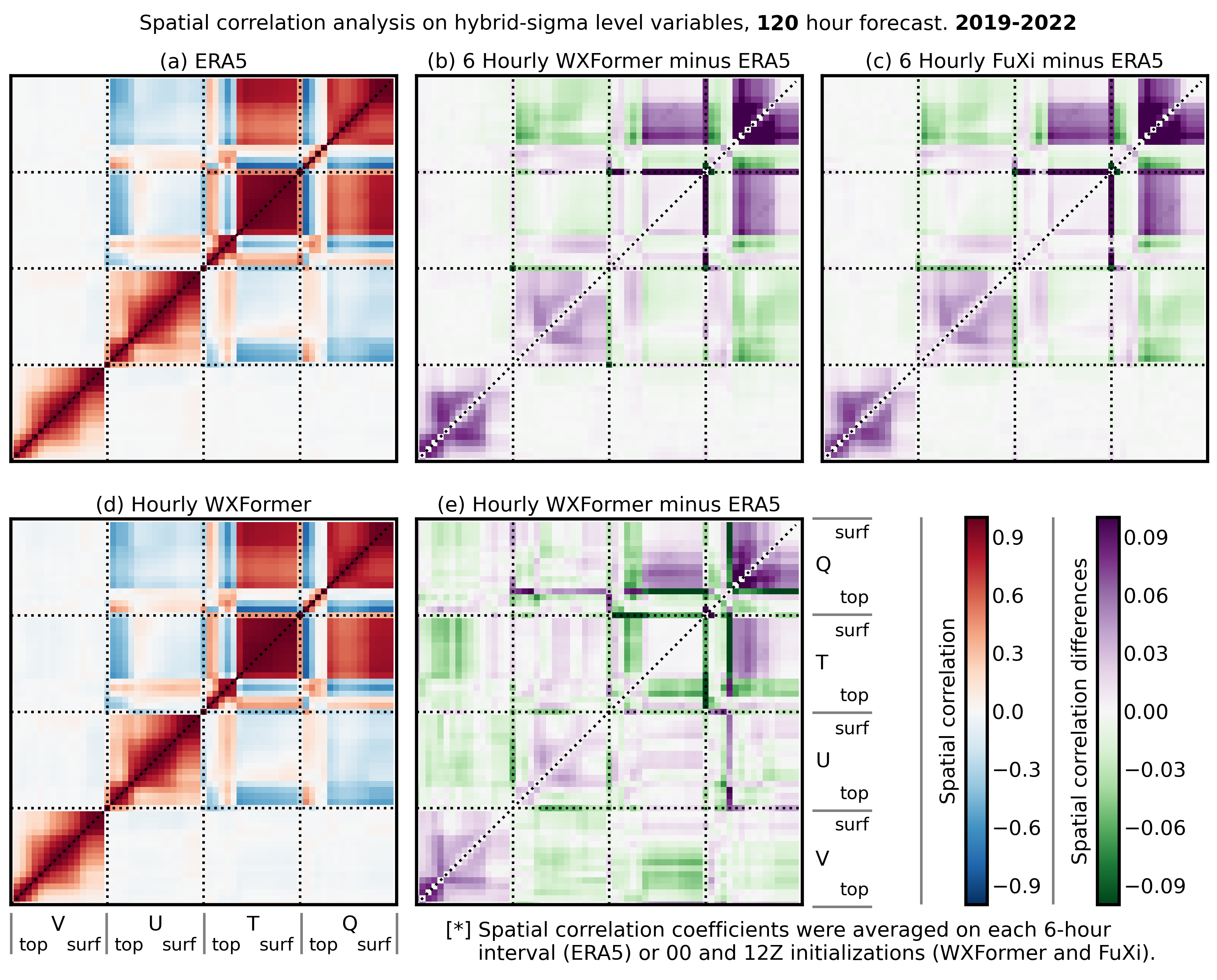}
    \caption{Spatial correlation analysis on model level meridional wind (``V''), zonal wind (``U''), air temperature (``T''), and specific humidity (``Q''); from model level top (``top'') to bottom (``surf''). (a) Self- and cross-correlation of ERA5 variables. (b) The difference of correlation coefficients between the 120-hour forecast of 6-hourly FuXi and ERA5 . (c) As in (b) but for 6-hourly WXFormer. (d) As in (a) but for hourly WXFormer. (e) As in (b), but for hourly WXFormer.}
    \label{fig4}
\end{figure}

Figure~\ref{fig4} illustrates the spatial correlation between upper-air diagnostic variables in WXFormer, FuXi, and ERA5 for a 120-hour forecast lead time. This analysis reveals that the AI-NWP models (WXFormer and FuXi) successfully retain the correct and coherent spatial relationships between key atmospheric variables even five days into the forecast, despite this coherence not being explicitly part of their training objectives.

The spatial correlation coefficients for both models generally exhibit the correct sign and magnitude in relation to concurrent upper-air variables, with deviations from ERA5 typically below 5\%. In panels (b) and (c), we observe that both WXFormer and FuXi slightly overestimate the positive correlation in specific humidity within the mid and lower troposphere, as well as the correlation between air temperature and specific humidity in the upper troposphere.

However, some challenges remain. The models struggle to capture the spatial coherence of meridional winds in the mid and upper troposphere, and correlations between surface-level temperature and humidity with upper-level variables are less accurately represented. Notably, FuXi appears to capture the relationships at the tropopause level more accurately than WXFormer, indicating slight model-specific variations in performance. The one-hourly model (e) at 120 struggles more than the 6-hourly models. This should be expected as the relative forecast skill at this period is relatively lower (Figure \ref{fig:one-shot}). Across most variables we see a growing incoherence between forecast fields, though some fields like the tropospheric specific humidity are much better represented.

\subsection{Case Study: Hurricane Laura (2020)}
\begin{figure}
    \centering
    \includegraphics[width=\linewidth]{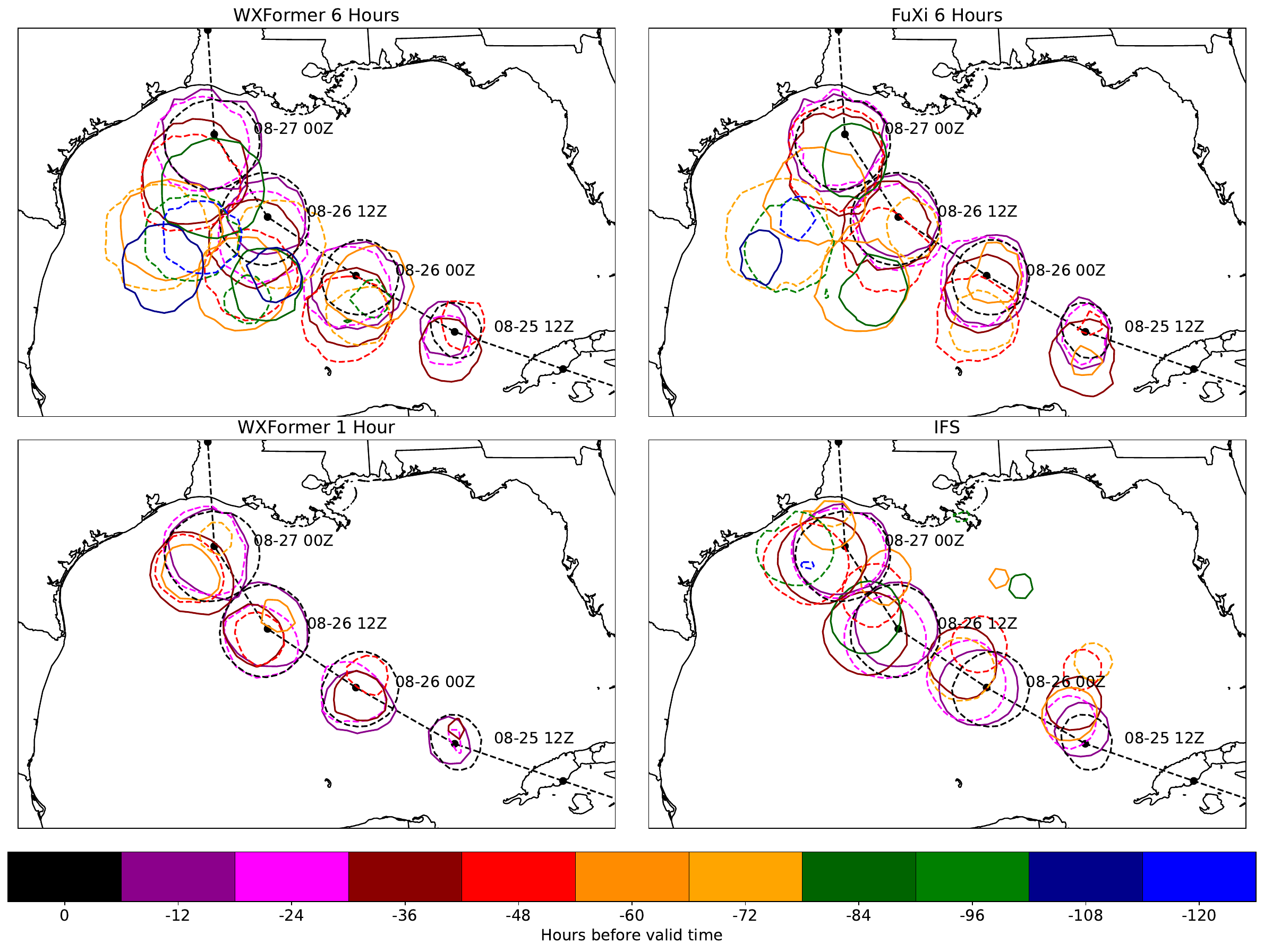}
    \caption{Comparison of Hurricane Laura 1000 hPa surface pressure predictions from 3 CREDIT AI NWP models and IFS. Each contour shows the prediction for a given valid time from different model initializations ranging from 12 to 120 hours in advance. Plotted track is from HURDAT. Dashed contours indicate predictions divisible by 24 hours.}
    \label{fig:laura}
\end{figure}

In addition to bulk verification statistics, CREDIT models are compared for the case of Hurricane Laura in 2020, which struck western Louisiana as a category 4 storm. In Fig.\ \ref{fig:laura}, the 1000 hPa surface pressure contour is used as a proxy for both track and intensity prediction with a larger contour indicating a more intense storm in each model. The 6-hour WXFormer model has the longest lead time on predicting the area of the 1000 hPa contour just before landfall with correct indications out to 120 hours (5 days). FuXi also had indications of the storm out to 5 days but with a weaker storm at day 5. Both had similar track errors and placed the storm too far south until 48 hours prior to landfall. The 1 hour WXFormer model had much lower track errors but had the storm much weaker prior to landfall. The IFS had track errors in the opposite direction of the AI models.

\section{Discussion}

\subsection{CREDIT as a model development framework}
The experimental results showcase the CREDIT framework's capability as a platform for AI NWP models. By successfully testing both an existing model (FuXi) and a new architecture (WXFormer), CREDIT demonstrates its ability to accommodate proven models while facilitating the development of innovative approaches. This flexibility is valuable in the field of weather prediction, where both building on established methods and exploring new techniques contribute to performance advancements.

The modification of FuXi within CREDIT highlights the potential advantages of working within a standardized framework. The addition of spectral normalization, padding (see supplemental), and non-negative correction on specific humidity simplified the original cascading architecture while maintaining or improving performance. This demonstrates how a well-designed framework can facilitate architectural improvements that might otherwise be challenging to implement and validate. The ability to rapidly test such modifications while maintaining reproducibility could prove beneficial for the research community.

Spectral analysis reveals that both AI models produce smoother forecasts over time, with reduced energy at smaller scales compared to IFS, while still maintaining competitive or better performance metrics. This finding suggests that there may be multiple viable approaches to handling atmospheric features across different scales. The AI models' behavior - losing fine-scale detail while maintaining forecast skill - warrants further study to better understand the relationship between spectral characteristics and forecast accuracy in different types of weather prediction systems.

\subsection{Insights from 6-hour model performance}

The trade-off between spectral fidelity and forecast accuracy revealed in this study is particularly intriguing. WXFormer and FuXi's loss of spectral power at longer lead times, resulting in smoother fields, may help prevent the propagation of small-scale errors, ultimately allowing for a more accurate representation of large-scale atmospheric features. Conversely, IFS, while producing more detailed forecasts, may include fine-scale inaccuracies that diverge from the true state of the atmosphere as forecast time progresses. This observation indicates that maintaining spectral fidelity is not always beneficial for long-term forecast accuracy, and highlights the need for a balanced approach in model development.

The impact of model architecture choices becomes apparent in the treatment of boundary conditions. The padding approach helped to address some of the artifact generation found at the polar and date-line regions, which are areas where many previous AI NWP models have faced difficulties. For models trained on a rectangular lat-lon grid, this approach offers a potential solution to the challenge of handling these geographic boundaries in global weather prediction. The implementation of this approach within CREDIT suggests that standardized solutions to common technical challenges could benefit researchers working in this field.


These results provide more evidence in support of further AI numerical weather prediction model development. The ability to both refine existing models and develop new ones within a single framework creates opportunities for much more rapid iteration and testing of new ideas compared with Fortran-based NWP models and even process-based emulators. As the field continues to explore the balance between traditional numerical methods and AI approaches, frameworks that support both innovation and standardization will play a role in advancing our understanding and prediction of atmospheric phenomena. The results demonstrated in this study pave the way for further exploration of AI-driven weather prediction models and underscore the potential value of standardized frameworks in atmospheric science research.

\subsection{Challenges of hourly AI forecasts}

In this study, we developed and trained the hourly WXFormer model using ERA5 data. However, its deterministic verification results did not reach the performance level of the 6-hourly models. Through our analysis, we identified several factors that may have contributed to the reduced accuracy in the hourly forecasts. One major takeaway from this work is the significant technical and resource challenges involved in moving from 6-hour to 1-hour timesteps.

High-frequency wave patterns emerged near the tropics and Southern Ocean in our day-2 forecasts and persisted throughout the 10-day forecast period (see Figure \ref{fig:waves} for an example). Additional anomalies were observed closer to the poles, including nonphysical spatial variations in some variables. The wave patterns were particularly pronounced in near-surface variables, such as 2-m air temperature and surface pressure, while their impact on 500 hPa geopotential height was muted. 

We intuit that the high-frequency waves and the small relative motion of the propagating patterns lead to resonance frequencies that are easily amplified by the model's learned dynamics. In contrast, the 6-hourly models benefit from inherent smoothing due to temporal averaging, which naturally mutes these waves. One attempted solution was to train the hourly model over a longer forecast roll-out period; however, this approach is computationally costly, introduces spatial smoothing (see Supplemental Figure \ref{fig:one-shot}), and negates some of the purpose of producing high-frequency output.

Initial experiments with running hourly predictions using 6-hourly pre-trained model weights showed promising results, with no high-frequency wave or other obvious anomalous patterns in the outputs. This suggests that the error pattern is unique to hourly models and appears unrelated to the model design. Transfer learning from 6-hourly models could potentially be a viable solution. 

We also experimented with different multi-step fine-tuning strategies, finding that while many of the models exhibited this high-frequency pattern, one model trained out to 120 hours with a modified multi-step loss to speed up the training, showed fewer wave patterns upon visual inspection. However, this model remained extremely cumbersome to train and did not achieve better overall performance than our primary hourly model shown here. While this suggests that the wave patterns might be addressable through training strategies, the computational costs and performance trade-offs require further investigation.

Lastly, we identified several technical operations that helped reduce the impact of high-frequency waves. Pole filtering and 2nd-order Laplacian-based diffusion filtering helps to dampen high-frequency waves by smoothing the model field. However, choosing when to apply the filter and to which variables remains an open question. Additionally, while boundary padding and non-negative corrections did not directly reduce high-frequency waves, they nonetheless contributed to improved deterministic verification scores for the hourly WXFormer model.

Future studies will be conducted to address the challenges above. The authors believe that CREDIT can be a powerful framework for investigating the error characteristics and forming systematic solutions of hourly AI forecasts.

\section{Conclusion}

This study demonstrates the effectiveness of the CREDIT framework through both the successful enhancement of the existing FuXi model and the development of the new WXFormer architecture. Both models achieve competitive or superior performance compared to operational forecasting systems, particularly at extended forecast ranges. The framework's ability to support model development while providing standardized training and evaluation protocols addresses a critical need in the atmospheric science community for reproducible, efficient AI-based weather prediction research.
 
Looking ahead, CREDIT provides a foundation for advancing AI-driven weather prediction through its combination of accessibility and scalability. As the field continues to evolve, the availability of a comprehensive, well-supported framework will enable researchers to focus on scientific innovation rather than technical implementation details. The successful results presented here suggest that CREDIT can serve as a catalyst for collaborative development in atmospheric modeling, potentially accelerating progress toward more accurate and computationally efficient weather forecasting systems.

\section*{Acknowledgments}
This material is based upon work supported by the NSF National Center for Atmospheric Research, which is a major facility sponsored by the U.S. National Science Foundation under Cooperative Agreement No. 1852977. This research has also been supported by NSF Grant No. RISE-2019758. We would like to acknowledge high-performance computing support from Derecho and Casper \cite{Cheyenne} provided by NCAR's Computational and Information Systems Laboratory, sponsored by the National Science Foundation. JSS thanks Jay Rothenberger, Daniel Howard, and Jared Baker for helpful conversations. DJG thanks Matthew Chantry and Jeff Anderson for helpful conversations and feedback. 

The ERA5 model-level reanalysis data for this study can be accessed through the NSF NCAR Research Data Archive at \url{https://www.doi.org/10.5065/XV5R-5344}. The neural networks described here and the simulation code used to train and test the models are archived at \url{https://github.com/NCAR/miles-credit}. The verification and data visualization code of this study is archived at \url{https://github.com/NCAR/CREDIT-arXiv}. Some of the code is being finalized with additional documentation and will be fully available at AGU24. For early access inquiries, please contact one of the authors.

\subsection*{Author Contributions}
JSS: project conception, developer of WXFormer, integration of NCAR-FuXi into CREDIT, multi-node multi-gpu scaling framework design, MPI workflow, spectral normalization, single- and multi-step training with gradient accumulation, model architecture and hyperparameter tuning, boundary padding, data workflow, loss functions, inference pipeline, visualization, software, writing.
YS: data pre-processing (z-score and residual normalization), software (data workflow for model training, single-step training routine, inference routine), model design (NCAR-FuXi, boundary padding, post-processing and correction), model training, result production and data saving, verification and quality control, data visualization, writing.
WEC: project conception, data gather, data packaging, data pre-processing and data ML model loading, software (data workflow, single-step training, normalization, inference pipeline, diffusion filtering, one-shot training, boundary padding, data pre-processing), visualization, analysis, writing.
DK: loss functions, model design discussions, software (data and program flow, distributed training, inference pipeline, data saving, python packaging and library compatibility, model schema unification).
JB: writing and funding 
SM: model and data workflow
AK: model design, single- and multi-step training, scaling pipeline, visualization, software
NS: multi-node multi-gpu framework, optimization of scaling framework, MPI workflow and NCCL backend optimization
BK: MPI workflow and environment optimization including optimized NCCL settings
DJG: project conception, data normalization, software (solar-forcing, boundary padding), data visualization, analysis, writing.

\appendix
\renewcommand{\thefigure}{A\arabic{figure}}
\setcounter{figure}{0} 

\section{Appendix}

\subsection{Padding Strategy}
\begin{figure}
    \centering
    \includegraphics[width=\linewidth]{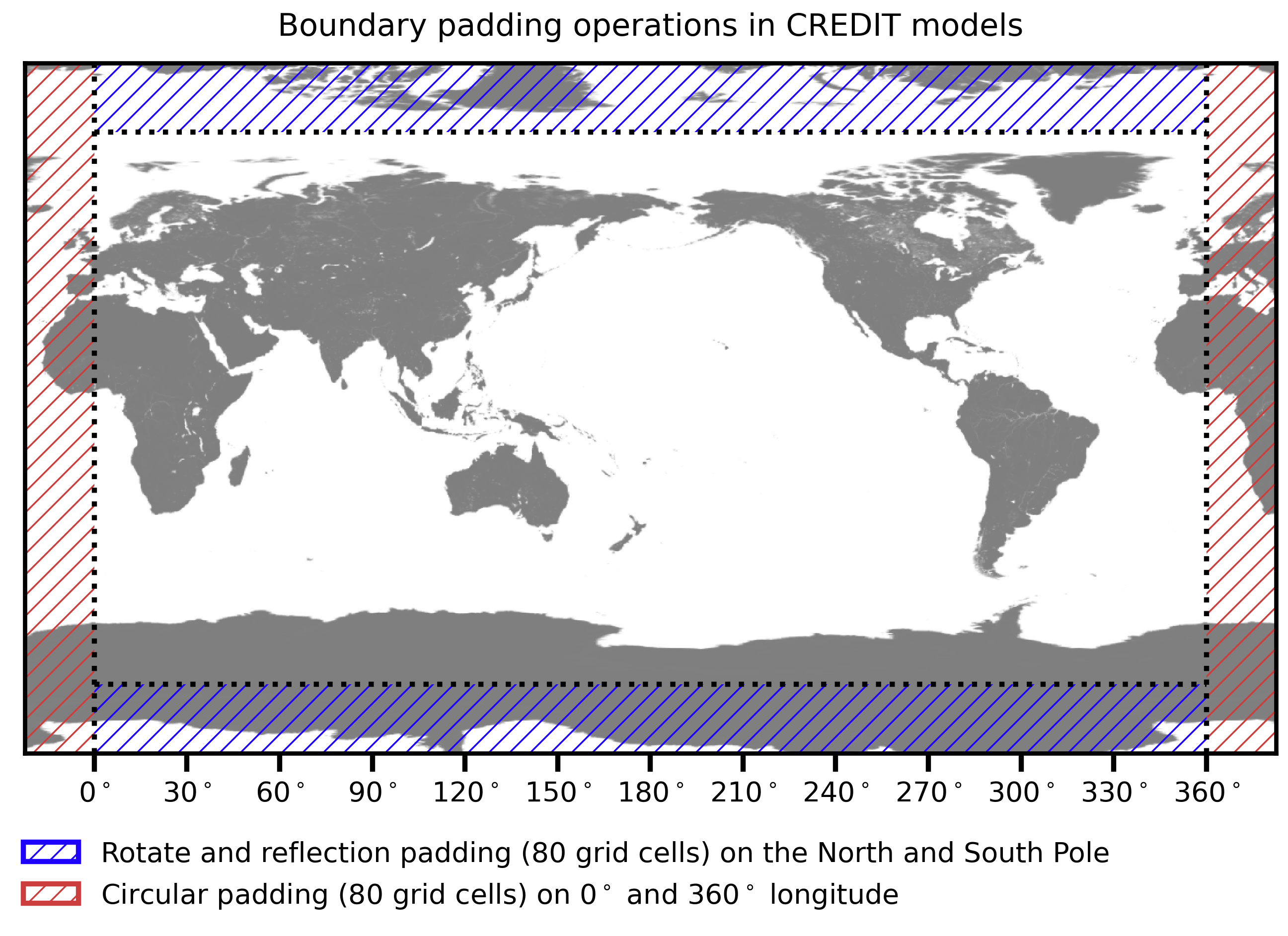}
    \caption{Boundary padding strategy used in CREDIT models WXFormer and FuXi, which demonstrates a periodic east-west boundary and rotated and reflected polar regions.}
    \label{fig:padding}
\end{figure}

In order to respect the unique conditions at the poles and east-west boundary, CREDIT applies the boundary conditions shown in Fig. \ref{fig:padding}. This padding approach includes a combination of rotated and reflected padding at the poles and circular padding at the longitudinal edges (0° and 360°), enabling a seamless, periodic east-west boundary and consistent handling of polar regions. The padding distance, in units of gridboxes, is set as a configuration parameter. For the FuXi and WXFormer models, the padding distance was set to 40 grid cells in the E/W and N/S direction. The rotate and reflection padding at the North and South Poles (blue shading) helps to mitigate artificial boundary effects that could arise from abrupt discontinuities. The N/S padding allows the ML model to achieve continuity across the pole and led to an improved total forecast score at every lead time in both WXFormer and FuXi compared to a simple reflected (N/S) padding in an ablation study. The circular padding along 0° and 360° longitude (red shading) supports periodic continuity in the model’s east-west direction, and prevented an artificial discontinuity from the ML models, which appeared when models were not given periodic information. Together, these padding strategies enhance the model’s ability to learn spatial relationships across global boundaries without introducing edge artifacts, supporting robust model performance in global-scale predictions. The N/S padding allows the ML model to achieve continuity across the pole and led to an improved total forecast score at every lead time in both WXFormer and FuXi compared to a simple reflected (N/S) padding. 

CREDIT allows the user to set multiple padding strategies and in future is exploring regridding to more physically consistent grid schema \cite{karlbauer2024advancing}, and supports models which are more grid agnostic \cite{Lam2023graphcast}, which will be demonstrated in a coming publication.

\subsection{Hourly WXFormer Multi-step Training}
\begin{figure}
    \centering
    \includegraphics[width=\linewidth]{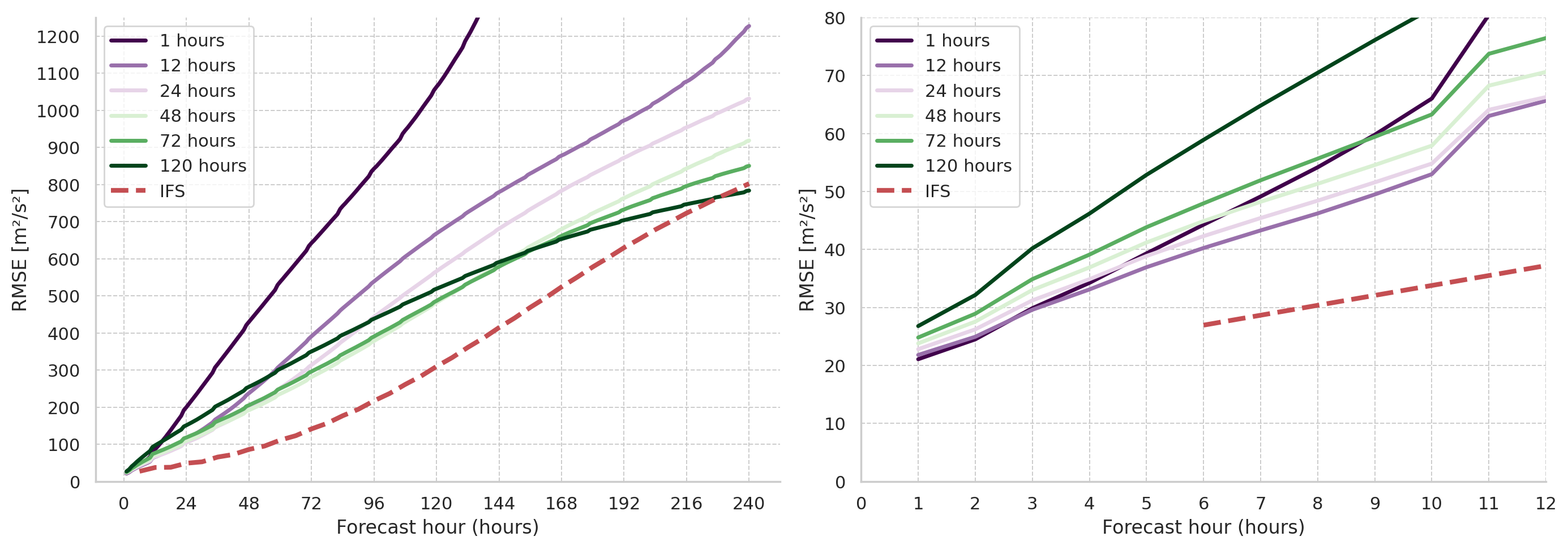}
    \caption{RMSE of Z500 forecasts for the hourly WXFormer model trained with varying one-shot lead-time targets. The legend indicates the lead time used during one-shot training, with each model evaluated on the 2020 validation period. (Left) Long-range RMSE performance up to 240 forecast hours. Models trained with longer one-shot targets (e.g., 120 hours, dark green) exhibit improved accuracy at extended lead times, demonstrating that longer training horizons reduce long-lead forecast error. (Right) Short-range RMSE performance up to 12 forecast hours. At shorter lead times, models trained with shorter one-shot targets (e.g., 1 hour, dark purple) perform best, indicating that shorter training intervals retain higher precision in early forecast hours. This trend suggests that longer lead-time training introduces temporal smoothing, benefiting long-range accuracy at the cost of short-term detail. These results highlight the need for tailored training strategies depending on forecasting priorities and computational resources.}
    \label{fig:one-shot}
\end{figure}

For this study, we explored the effects of what we term as ``one-shot'' training on the forecast accuracy of the WXFormer model. One-shot training refers to a method in which the model is trained to predict a specific lead time, calculating the loss only at that designated forecast horizon. This approach contrasts with traditional sequential training methods, as it focuses exclusively on optimizing model performance for the chosen lead time, and saves on the computational cost of accumulating gradients. Note this is different than the strategy described in section \ref{model_training}. Note also the residual loss was not yet implemented in the CREDIT framework with these models, potentially hindering performance improvement. Figure \ref{fig:one-shot} illustrates the RMSE of the Z500 field for models trained with one-shot targets ranging from 1 hour to 120 hours at 12 and 24-hour intervals. The results are then tested on the 2020 forecast validation period, as described in the main text.

The left panel displays the RMSE across various lead times up to 240 hours, while the right panel zooms in on short lead times (up to 12 hours). As shown in the left panel, increasing the one-shot training lead time improves long-lead forecast performance, with the RMSE decreasing for models trained at progressively longer horizons. For example, the model trained with a 120-hour one-shot target (represented by the dark green) exhibits significantly lower RMSE at 240 hours than models trained with shorter lead times. This result suggests that extended one-shot training better equips the model for long-lead forecasts, making it particularly useful for subseasonal-to-seasonal (S2S) applications. However, as discussed in the main paper, training a 1-hour model over such long lead times is computationally intensive, requiring substantial core hours and resources, and presents practical challenges.

In contrast, the right panel reveals an inverse trend at shorter lead times. If the goal is to maximize model skill for short-term forecasts, it is more effective to train the model with shorter lead-time targets. For example, the 1-hour trained model (dark purple line) outperforms models trained with longer lead times at very short horizons, with the green line (120-hour training) showing the highest RMSE at the 1-hour mark. This result implies that training with longer lead times introduces a degree of temporal smoothing in the model, which benefits longer forecasts but sacrifices precision at short lead times. We note, that the models trained out to 72 and 120 hours did not display the nonphysical wave patterns shown in figure \ref{fig:waves}. 

Overall, this analysis highlights the need to balance model smoothness, computational resources, and the specific forecasting goals. For S2S applications, training with extended one-shot lead times may provide enhanced long-lead forecast accuracy, but at a substantial computational cost. In contrast, for applications that prioritize short-term precision, shorter one-shot training intervals are more appropriate. While this study was conducted on a subset of model levels as part of an ablation analysis, the insights here should be valuable to the forecasting community in optimizing model training strategies based on intended use cases. Additionally, this highlights the use case of CREDIT as a platform to explore ideas of model training and optimal ML based forecasting techniques. 

\subsection{Hourly WXFormer Instabilities}

\begin{figure}
    \centering
    \includegraphics[width=\linewidth]{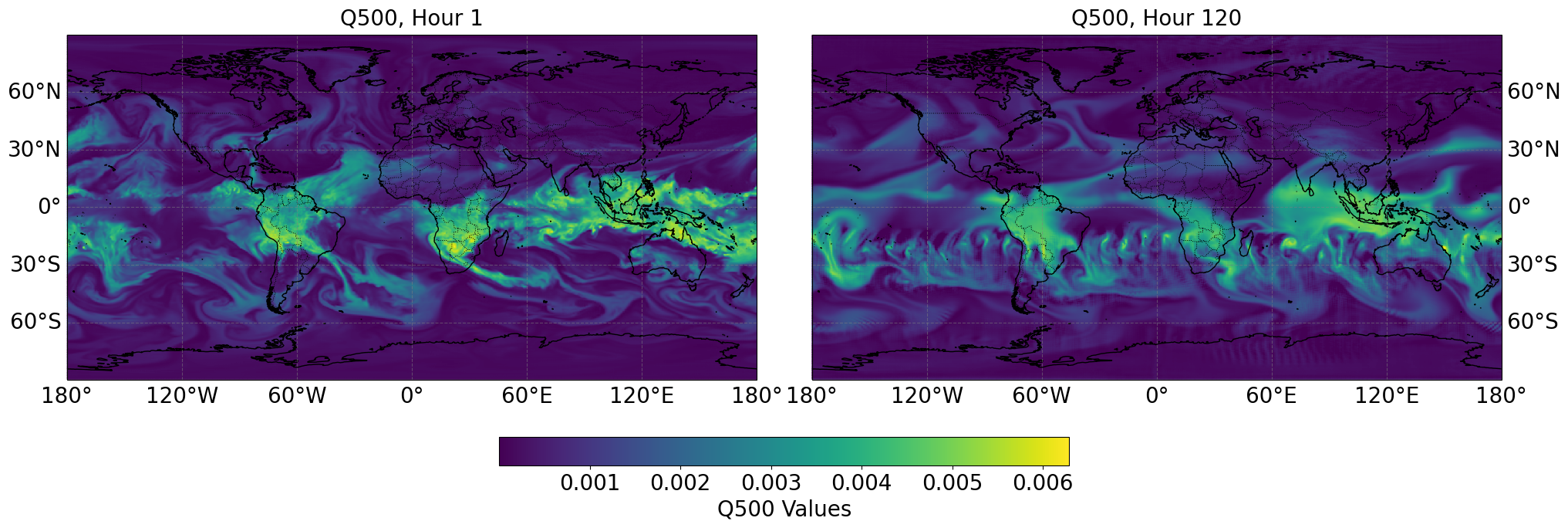}
    \caption{Waves appearing in the southern hemisphere at 120 hours lead time on 500 hPa specific humidity (``Q500'') using the 1-hour WXformer model.}
    \label{fig:waves}
\end{figure}

Figure~\ref{fig:waves} illustrates non-physical wave patterns that sometimes develop in the WXFormer model’s Q500 field after 5 days (120 hours) of continuous hourly roll-out. In the left panel, representing 1-hour model forecast, the specific humidity field shows a physically realistic representation of mid-tropospheric specific humidity. However, by hour 120 (right panel), spurious wave-like artifacts become evident in the Southern Hemisphere. These patterns likely arise from instability in the model dynamics during extended lead times, suggesting that additional training adjustments or post-processing techniques may be needed to improve long-range forecast stability and accuracy in the WXFormer model for high temporal resolution simulations.

\bibliographystyle{unsrtnat}
\bibliography{creditArxiv}

\end{document}